\documentclass[letterpaper, 10 pt, conference]{ieeeconf}  

\IEEEoverridecommandlockouts                              

\usepackage{amsmath} 
\usepackage{color}
\usepackage{import}
\usepackage[bookmarks=true,hidelinks]{hyperref}
\usepackage{siunitx}
\usepackage{subfig}
\usepackage{booktabs}
\usepackage{graphicx}
\usepackage{ifthen} 

\title{\LARGE \bf
DigiArm: An Anthropomorphic 3D-Printed Prosthetic Hand with Enhanced Dexterity for Typing Tasks
}

\newboolean{isARXIVversion}

\setboolean{isARXIVversion}{true}

\ifthenelse{\boolean{isARXIVversion}}{
\author{Dean Zadok$^{1}$, Tom Naamani$^{2}$, Yuval Bar-Ratson$^{2}$, Elisha Barash$^{2}$, \\ Oren Salzman$^{1}$, Alon Wolf$^{2}$, Alex M. Bronstein$^{1}$ and Nili Krausz$^{2}$
\thanks{$^{1}$Department of Computer Science, Technion, Haifa, Israel
        {\tt\small \{deanzadok,osalzman,bron\}@cs.technion.ac.il}}%
\thanks{$^{2}$\mbox{Department of Mechanical Engineering, Technion, Haifa, Israel}
        {\tt\small \{tom.naamani,yuval.bar,}\newline
        {\tt\small elishabarash\}}%
        {\tt\small @campus.technion.ac.il}\newline
        {\tt\small \{alonw,nili.krausz\}@me.technion.ac.il}}%
}}
{\author{Anonymous Authors
\thanks{Paper under double-blind review.}}}

\newcommand{\Cpp}{C\raise.08ex\hbox{\tt ++}}

\newcommand{\ignore}[1]{}

\newboolean{HIDENOTES}
\setboolean{HIDENOTES}{false}

\ifthenelse{\boolean{HIDENOTES}}{
    \newcommand{\OS}[1]{{}}
    \newcommand{\DZ}[1]{{}}
    \newcommand{\AB}[1]{{}}
    \newcommand{\AW}[1]{{}}
    \newcommand{\CONT}[1]{{}}
}{
    \newcommand{\OS}[1]{\textcolor{red}{#1}}
    \newcommand{\DZ}[1]{\textcolor{blue}{#1}}    
    \newcommand{\AB}[1]{{\textcolor{green}{#1}}}    
    \newcommand{\AW}[1]{{\textcolor{yellow}{#1}}}

}

\begin{document}

\maketitle
\thispagestyle{empty}
\pagestyle{empty}

\begin{abstract}

Despite recent advancements, existing prosthetic limbs are unable to replicate the dexterity and intuitive control of the human hand.
Current control systems for prosthetic hands are often limited to grasping, and commercial prosthetic hands lack the precision needed for dexterous manipulation or applications that require fine finger motions.
Thus, there is a critical need for accessible and replicable prosthetic designs that enable individuals to interact with electronic devices and perform precise finger pressing, such as keyboard typing or piano playing, while preserving current prosthetic capabilities.
This paper presents a low-cost, lightweight, 3D-printed robotic prosthetic hand, specifically engineered for enhanced dexterity with electronic devices such as a computer keyboard or piano, as well as general object manipulation.
The robotic hand features a mechanism to adjust finger abduction/adduction spacing, a 2-D wrist with the inclusion of controlled ulnar/radial deviation optimized for typing, and control of independent finger pressing.
We conducted a study to demonstrate how participants can use the robotic hand to perform keyboard typing and piano playing in real time, with different levels of finger and wrist motion.
This supports the notion that our proposed design can allow for the execution of key typing motions more effectively than before, aiming to enhance the functionality of prosthetic hands.

\end{abstract}



\section{INTRODUCTION}



%

%

%
Existing robotic prosthetic arms have revolutionized the lives of amputees, enabling many basic daily functions. 
Current prosthetic hands predominantly rely on myoelectric interfaces to achieve multiple grasping behaviors ~\cite{herberts1969myoelectric,simao2019emg,DBLP:journals/corr/abs-1801-07756}, but the desire to provide fine motor control has led researchers to develop interfaces that can differentiate between intended finger motions and provide proportional control~\cite{simpetru2023proportional,li2023simultaneous,ZadokSWB23,zadok2025inferring}.
The current state-of-the-art multi-joint prosthetic hand designs include both anthropomorphic~\cite{butterfass2001dlr,huang2006mechanical,resnik2014deka,lenzi2016ric,yang2025lightweight,DBLP:journals/access/RomeroGPCSLRAE25,controzzi2016sssa} and non-anthropomorphic~\cite{dollar2010highly,deimel2013compliant,manti2015bioinspired} architectures that aim to replicate human dexterity.

However, despite these technological advances, existing solutions face a critical limitation: they struggle to balance affordability with the precision required for tasks involving intricate finger motions such as keyboard typing or piano playing.
Commercial prosthetic hands capable of fine manipulation (such as the LUKE arm \cite{resnik2014deka}) remain expensive and in limited use, while lower-cost alternatives typically concentrate on essential daily activities.
Moreover, the design of prosthetic hands often prioritize predetermined grasps with strong, stable grips for holding objects, which can come at the cost of reduced fine motor skills essential for interactions with electronic devices, such as in tasks that require precise and independent finger movements like typing on a keyboard.

\begin{figure}[t]
\centering
\includegraphics[trim=200 150 150 150, clip, width=0.425\textwidth]{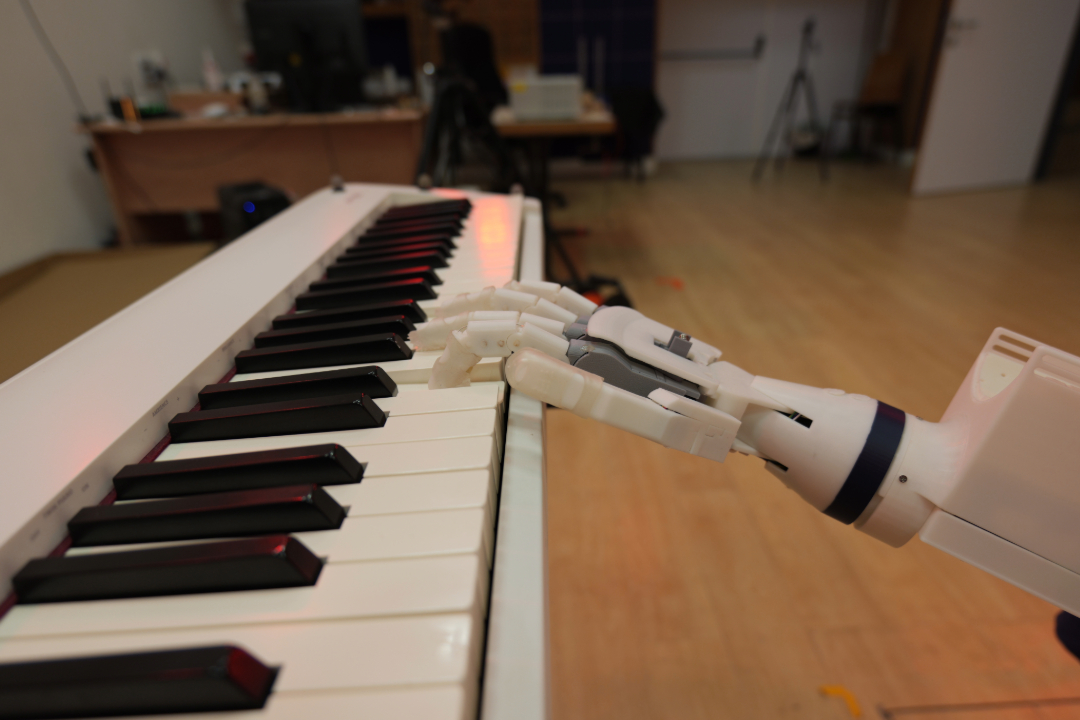}
\vspace{-0.1cm}
\caption{The developed robotic prosthetic hand. The system features independent finger control, adjustable finger spacing, and a 2D wrist optimized for keyboard typing and piano playing applications. All components are 3D-printed with affordable electronics and actuators.}
\label{fig:robotic_hand}
\vspace{-0.4cm}
\end{figure}

%


This paper presents a novel anthropomorphic, low-cost, and 3D printed robotic hand specifically engineered to bridge the gap between affordability and dexterity for fine finger motions.\ifthenelse{\boolean{isARXIVversion}}{\footnote{Implementation instructions are at \textcolor{blue}{\href{https://github.com/deanzadok/digiarm}{https://github.com/deanzadok/digiarm}}.}}{\footnote{Our implementation and installation instructions will be publicly available upon publication.}}
Our research makes three primary contributions: 
(i) \emph{A robotic hand design} that can execute fine finger pressing motions for keyboard typing and piano playing while maintaining anthropomorphic form and essential grasping capabilities. The design features independent finger control, adjustable finger spacing, and optimized wrist motion for extended reachability.
(ii) \emph{An open-source prosthetic platform} with complete accessibility for personal use. The platform includes 3D-printable components, commercially available motors, and inexpensive controllers, enabling widespread adoption by users and researchers.
(iii)~\emph{Empirical insights from human experiments} while performing keyboard typing and piano playing tasks, providing valuable insights regarding the operation of prosthetic hands in these applications. These lessons underscore the importance of finger spacing adjustments and wrist motion in prosthetic design, for tasks that require multi-digit coordination.



\section{RELATED WORK}


Over the past two decades, significant progress in robotic technology has led to remarkable developments in upper-limb prosthetics. In the aftermath of the wars in Iraq and Afghanistan, the DARPA Revolutionizing Prosthetics program~\cite{burck2011revolutionizing,larson2023role} spurred rapid developments and new and exciting areas of research both in mechatronic development as well as controls and sensory feedback for these devices. 
A key goal of the project was the development of a prosthetic arm that could replicate the degrees of freedom of the human arm (this project eventually led to the commercialization of the DEKA LUKE arm~\cite{resnik2014deka}). 
During the same years, various prosthetic hand prototypes were developed in  academia~\cite{lenzi2016ric,controzzi2016sssa,weiner2018kit,della2015dexterity,cipriani2011smarthand} with multi-articulating architectures enabling multiple grasp types. 

Subsequently, various multi-joint prosthetic hands have been commercialized, including TASKA \cite{taska}, COVVI \cite{covvi}, Mia \cite{mia_hand}, Michelangelo~\cite{luchetti2015impact}, AbilityHand \cite{psyonichand}, and the iLimb~\cite{van2013functionality}. Most feature five or six degrees of freedom (DOFs) with single-DOF underactuated fingers, and in general a single DOF wrist (supination/pronation), with a few examples of two-DOF wrists with the addition of flexion/extension (Michelangelo \cite{michelangelohand}, Hero Pro hand \cite{heroprohand}). 
Importantly, democratization of access to robotic hands has gained traction, and multiple open-source platforms exist that enable continuing research in the field~\cite{DBLP:journals/access/RomeroGPCSLRAE25,ma2013modular,said2019customizable,krausz2015design}.

Current prosthetic devices are primarily controlled using electromyography (EMG) ~\cite{herberts1969myoelectric,DBLP:journals/corr/abs-1801-07756,simao2019emg,simpetru2023proportional,englehart2003robust}
but in recent years, additional control modalities to improve the functionality of prosthetic hands were introduced. 
These include ultrasound imaging~\cite{ZadokSWB23,zadok2025inferring,DBLP:conf/chi/McIntoshMFP17,DBLP:journals/tsmc/YangCHYL21}, magnetomyography~\cite{greco2023discrimination,gherardini2024restoration}, EEG~\cite{maibam2024enhancing}, and even motion-based approaches~\cite{uyanik2019deep,DBLP:journals/corr/abs-2503-17846}.
These sensing modalities enable data-driven approaches that extract information from user intent gestures.
Popular examples include detecting prominent hand gestures or grasp patterns~\cite{simao2019emg,DBLP:journals/corr/abs-1801-07756,DBLP:conf/chi/McIntoshMFP17}
, and learning proportional control to map signals to continuous joint motions~\cite{simpetru2023proportional, li2023simultaneous, ZadokSWB23,zadok2025inferring,nowak2023simultaneous}.
These algorithmic advances enable prosthetic arms to perform specialized tasks, such as playing musical instruments~\cite{zadok2025inferring,4696889}, engaging with electronic devices~\cite{DBLP:conf/chi/HuWGYS21,chen2025robotic}, and employing modular designs that can switch end effectors for specific applications~\cite{chappell2025beyond}.
While these advances have expanded prosthetic capabilities, there remains room for exploring mechanical-design approaches specifically tailored for tasks that require fine finger motions such as keyboard typing and piano playing.

\section{SYSTEM DESIGN}
\label{sec:sys-design}

\begin{figure*}[t]
\centering
\includegraphics[trim=0 0 0 0, clip, width=0.8\textwidth]{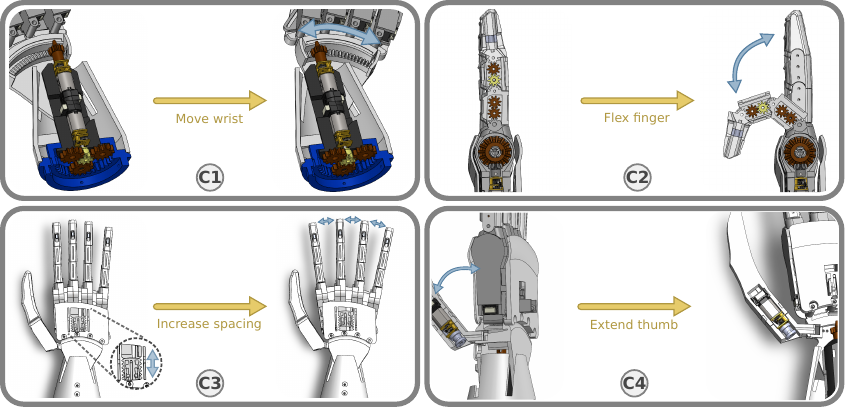}
\vspace{-0.1cm}
\caption{
Technical snapshots of the main mechanisms of the DigiArm, with each block demonstrating the mechanism in two different states.
(\textbf{C1}) An internal view of the 2D-wrist mechanism: ulnar-radial deviation is powered by the top motor, and pronation-supination is powered by the bottom motor.
(\textbf{C2}) Assembled finger module showing the motor base housing, bevel gear system, and cable-driven actuation mechanism connecting to the distal phalanx. 
(\textbf{C3}) The ``splay'' mechanism for adjustable finger spacing showing the finger spacing while closed and opened.
(\textbf{C4}) The Thumb module illustrating the integrated motor design and bidirectional rotational motion from extension to flexion.
}
\label{fig:solid}
\vspace{-0.4cm}
\end{figure*}

\subsection{Overview}
\label{ssec:overview}


Our robotic hand platform, DigiArm, was developed with four primary design objectives to ensure the system meets both technical performance requirements and practical needs:

\begin{itemize}
    \item[\textbf{O1}] Functionality---achieving high-precision control for delicate key pressing without sacrificing performance in object grasping tasks.
    \item[\textbf{O2}] Mobility--- lightweight portability for daily use.
    \item[\textbf{O3}] Affordability---allowing for widespread adoption by maintaining a low-cost design.
    \item[\textbf{O4}] Easy Integration---providing an easy-to-use communication interface for real-time control.
\end{itemize}

Functionality (\textbf{O1}) represents the core technical requirement, specifically optimized for keyboard-typing and piano-playing applications. The system must provide precise position control for accurate key targeting without unwanted adjacent key presses, and controllable speed regulation for different typing styles and musical tempos.

Mobility (\textbf{O2}) requires that the complete system, including the power source, approximate the weight of a natural arm to prevent fatigue and maintain natural movements during extended use. This constraint directly influences component and battery selection.

Affordability (\textbf{O3}) addresses the cost barrier in prosthetic adoption by providing an open-source platform with readily available components, enabling cost-effective manufacturing and allowing researchers to modify and extend the design while keeping it accessible.

Easy integration (\textbf{O4}) ensures that sophisticated actuation options remain accessible through an intuitive interface. The real-time controller translates high-level commands (e.g., \texttt{flex finger $F2$ by $50$ degrees}) into coordinated motor movements via standardized protocols, allowing the controller to interface with various input methods, whether neuromuscular sensors or autonomous control systems.


\subsection{Robotic Hand Mechanical Design}
\label{ssec:robotic-hand-mechanical-design}
 Our hand design consists of four major components:

\begin{itemize}
    \item[\textbf{C1}] A two DOF \textbf{wrist} capable of performing pronation/supination and ulnar/radial motions, designed to hold the five fingers securely during operation.
    \item[\textbf{C2}] Four underactuated \textbf{finger} modules, each consisting of three links actuated using a cable mechanism.
    \item[\textbf{C3}] A manual adjustment mechanism we call \textbf{``Splay''}, that modifies by controlling the spacing between adjacent fingers (aka adduction/abduction).
    \item[\textbf{C4}] A \textbf{thumb} with a single DOF designed for pressing keys while preserving opposition for grasping and pinching.
\end{itemize}

When assembled, these components form an anthropomorphic robotic hand optimized for piano and keyboard operation while maintaining object manipulation capabilities similar to previous prosthetic designs.

The 2D-wrist component (\textbf{C1}) enables two essential motions for dexterous manipulation (see Fig.~\ref{fig:solid}, \textbf{C1}). First, pronation/supination allows the hand to orient for different grasping configurations and provides the positioning necessary for certain tasks, such as turning objects or adjusting grip orientation. This motion provides a total range of \SI{230}{\degree}, with \SI{190}{\degree} of supination and \SI{40}{\degree} of pronation from a neutral palm-down position.
Second, ulnar/radial deviation addresses a critical limitation in existing prosthetics by enabling natural wrist positioning during typing and piano playing. This mechanism provides \SI{60}{\degree} of total range of motion, symmetric about the sagittal plane (\SI{30}{\degree} in each direction from the center), and allows for reaching a wide range of keys without excessive arm movement, whether accessing distant keyboard keys or spanning octaves on a piano. By providing this, the system is designed to reduce the compensatory shoulder and elbow movements that users typically employ with prosthetic wrists lacking sufficient DOFs, thereby improving comfort and reducing fatigue during extended use.

The ulnar/radial deviation is actuated by a $12$:$132$ bevel gear pair, with the smaller gear mounted to the forward motor within the tower (the component shown in black in Fig.~\ref{fig:solid}, \textbf{C1}).
The larger gear is printed in-place as part of the bridge part, which mounts directly onto the tower, ensuring precise geometric meshing constraints. The gear positioning on the upper portion of the bridge (when palm faces downward) allows gravity to assist with gear engagement. Pronation/supination is actuated using the rear motor of the tower, which drives a $9$-tooth sun gear in a planetary gear system with a $9$:$13$:$36$ ratio. This system incorporates three $13$-tooth planet gears and a constrained $36$-tooth outer ring that is fixed by the socket assembly.
For force values at the wrist and the other joints, please refer to Sec.~\ref{sec:app}.



For simplicity, each of the four fingers (\textbf{C2}) features an identical design composed of three phalanges and a motor-base housing. The motor, a \SI{6}{\volt} DC motor within the base, incorporates a $1$:$75$ gear ratio and drives a bevel gear that actuates a fishing line cable connected to the distal phalanx. The selected gear ratio balances fast motion with the torque required for pressing keyboard keys and piano notes. The relative motion between phalanges is constrained by an embedded gear train within the phalanges, which ensures coordinated flexion movement during operation (see~Fig.~\ref{fig:solid},~\textbf{C2}). The distal phalanx achieves a flexion range from \SI{180}{\degree} in full extension to \SI{15}{\degree} when fully closed, providing \SI{165}{\degree} of total motion.
To maintain consistent cable tension after prolonged use, the distal phalanx has a screw-based tensioning mechanism. This mechanism can be manually adjusted to tighten the cable when slack develops, ensuring reliable actuation throughout the operational lifetime of the device.


The ``Splay'' mechanism (\textbf{C3}) enables manual adjustment of finger spacing to simulate natural adduction/abduction movements of fingers. This allows users to modify the distance between all four fingers simultaneously by operating the mechanism with their contralateral hand.
The mechanism operates through a movable rack-and-pinion system, where a single rack with four slots engages with individual pinions attached to each finger base. By sliding the rack linearly within a fixed cover, users can incrementally adjust finger spacing across five discrete positions and lock the configuration in place (see Fig.~\ref{fig:solid}, \textbf{C3}). The spacing ranges from a fully closed configuration, where all fingers are parallel, to a fully open configuration with angles of the Index, Middle, Ring and Little fingers relative to the central axis of
  at \SI{-12}{\degree}, \SI{0}{\degree}, \SI{10}{\degree}, and  \SI{19}{\degree}, respectively.
This enables the customization of the hand span for different keyboard layouts or musical chords without requiring additional motors or electronic control.

%

The thumb module (\textbf{C4}) features a simplified single DOF design optimized for both key pressing and object manipulation. The thumb is positioned at a \SI{40}{\degree} angle relative to the main axis of the hand, with rotational motion (see Fig.~\ref{fig:solid},~\textbf{C4}) specifically configured to enable effective key-pressing gestures while preserving essential grasping capabilities, including pinching objects in coordination with the index finger and providing support for holding larger objects (see Fig.~\ref{fig:objects_grasp}).
Unlike the four finger-modules, the thumb actuation system integrates a \SI{6}{\volt} DC motor with a $1$:$100$ gear ratio directly within the thumb link itself. This higher gear ratio reduces the actuation speed, which was found to be excessive with the original configuration.
The motor drives a tiny spool mounted on the motor shaft, with a fishing line cable wound around the spool to enable bidirectional rotational motion of the thumb according to the shaft rotation.
The thumb achieves a total range of motion from \SI{-10}{\degree} (extended slightly beyond palm level) to \SI{90}{\degree} (fully flexed), providing a total of \SI{100}{\degree} of rotation. This integrated design maintains a compact profile while eliminating the need for additional motor base housing, thereby reducing the overall system weight and complexity.


\subsection{Closed-loop Control}
\label{ssec:closed-loop_control}

\begin{figure}[t]
\centering
\includegraphics[trim=0 0 0 0, clip, width = 0.425\textwidth]{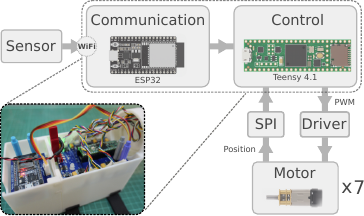}
\vspace{-0.1cm}
\caption{
DigiArm control system architecture.
The ESP32-based communication module receives sensor commands via WiFi and transmits them to the Teensy 4.1, which performs real-time closed-loop control of seven motors using PWM signals and SPI-based communication for encoder updates.
}
\label{fig:control_diagram}
\vspace{-0.4cm}
\end{figure}


To operate the seven DOFs of the DigiArm, the system requires seven miniature high-power carbon brush DC motors, each connected through a 6-pin connector that delivers pulse-width modulation (PWM) signals, direction control, and encoder communication.
The controller architecture consists of two main components: an ESP32-based circuit that manages communication with input sensors, and a Teensy 4.1-based circuit with SPI communication and seven dedicated connectors for simultaneous motor and encoder control. The system operates at \SI{7.4}{\volt} using two rechargeable batteries. 

Since the motors utilize incremental rather than absolute encoders, the system initiates with a calibration process where each motor moves through its full range of motion to detect position boundaries.
The controller receives messages containing three elements: a byte with bit-flags indicating which motors to actuate, an array of normalized encoder positions ($0$-$255$ range) representing target positions relative to calibrated limits, and corresponding PWM values specifying motion speeds.
Upon message reception, the controller interprets commands, reads encoder values, calculates required distances and directions, and executes coordinated multi-motor movements. Complete finger flexion requires approximately \SI{0.4}{\second}, with communication latency of \SI{0.05}{\second} to \SI{0.1}{\second} for wireless transmission depending on signal quality (negligible when wired). Motion continues until target positions are reached or stall detection occurs through buffer-based encoder monitoring. The controller manages simultaneous motor operations through iterative monitoring during execution.

\section{EVALUATION}
\label{sec:evaluation}

\begin{figure}[t]
\centering
\subfloat[\label{fig:key_press_force_setup}]{\includegraphics[trim=40 40 190 0, clip, width=0.176\textwidth]{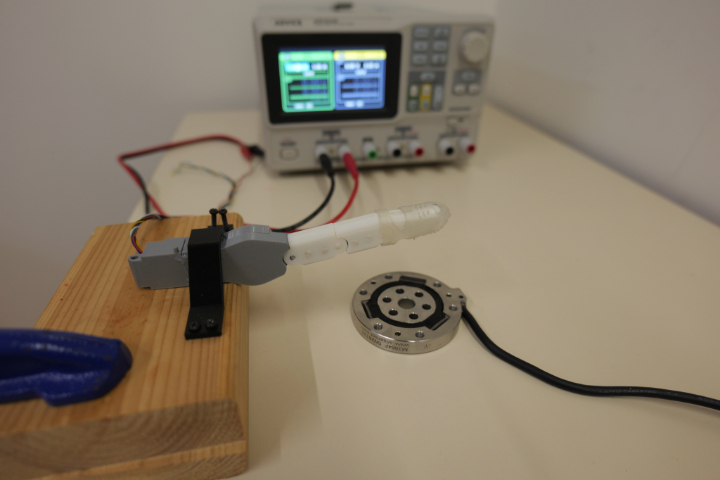}}\hspace{1.5mm}%
\subfloat[\label{fig:key_press_force_results}]{\includegraphics[trim=0 0 0 0, clip, width=0.264\textwidth]{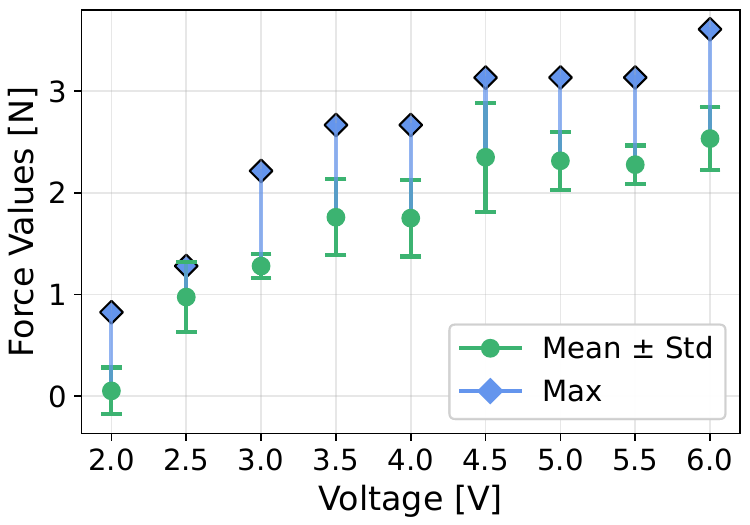}}
\vspace{-0.1cm}
\caption{
Force measurement experiment.
(a) Our experimental setup showing a finger module positioned above a 6-axis force sensor for quantifying key-press forces.
(b) Force output as a function of input voltage, displaying mean with standard deviation (green) for the average force during the pressing duration and maximum values (blue) during contact events, with a maximum output of \SI{3.6}{\newton} at \SI{6}{\volt} input.
}
\vspace{-0.5cm}
\end{figure}

\begin{figure*}[t]
\centering
\subfloat[\label{fig:piano_heatmap}]{\includegraphics[trim=75 0 38 0, clip, width=0.525\textwidth]{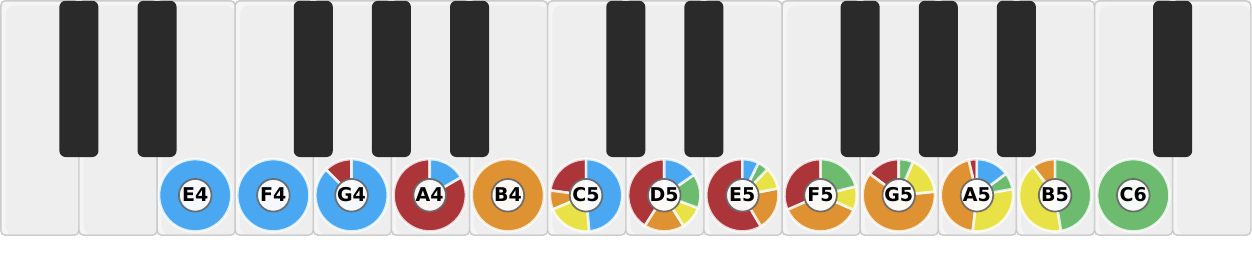}}\hspace{5mm}%
\subfloat[\label{fig:keyboard_heatmap}]{\includegraphics[trim=0 0 0 0, clip, width=0.425\textwidth]{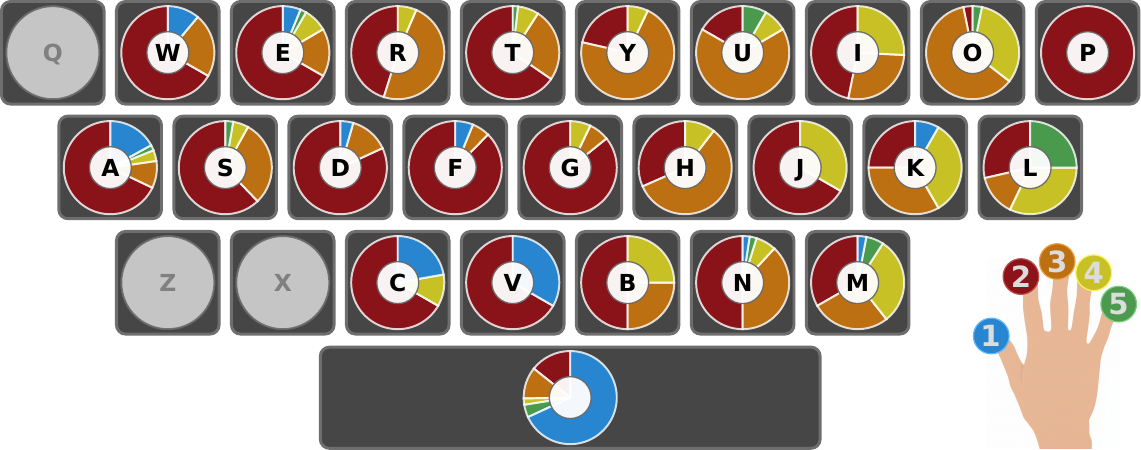}}
\vspace{-0.2cm}
\caption{
Finger usage patterns during key-pressing tasks.
(a) Piano key selection heatmap showing correspondence patterns between finger choice and key position, with a slight preference for thumb and finger gestures.
(b) Keyboard typing heatmap revealing index finger dominance across most keys, with an interesting shift towards outer fingers (middle, ring, little) correlated with rightmost key positions, and clear thumb specialization for the space key. Results demonstrate consistent finger selection strategies across subjects during prosthetic hand operation.
}
\vspace{-0.5cm}
\end{figure*}

To evaluate the DigiArm and its control architecture, we designed an interface to control the hand using motion capture (Sec.~\ref{ssec:system_integration}).
Our validation consists of evaluating finger force output for actuation (Sec.~\ref{ssec:key_press_force_requirements}), conducting a study with eleven users performing typing and piano tasks, and evaluating how adjustable finger spacing and wrist motion affect performance (Sec.~\ref{ssec:user_study}),
and demonstrating preserved grasping capabilities for everyday objects (Sec.~\ref{ssec:preserving_grasping_capabilities}).

\subsection{System Integration}
\label{ssec:system_integration}

For experimental implementation, we developed a complete wearable system with multiple components.
The controller is housed in a custom ``bypass'' socket with integrated straps, allowing able-bodied users to wear the device beneath their arm for motion-controlled operation during key-pressing tasks (see Fig.~\ref{fig:pressing_examples}). We also designed a socket configuration for individuals with below-elbow amputations, though experiments were conducted with able-bodied users.
Power is provided by two \SI{3.7}{\volt} batteries (\SI{165}{\gram}) worn on the upper arm. The DigiArm weighs \SI{341}{\gram} including motors and cables, while the complete system weighs \SI{648}{\gram}. Total cost including electronics is approximately \$250, with potential reduction through lower-cost motors. Assembly instructions and specifications will be provided upon publication.

To evaluate the integrated system of the DigiArm without unnecessary influence on the part of the chosen sensor modality or intent prediction algorithm (e.g. EMG-based pattern recognition vs. ultrasound-based joint mappings) we implemented a motion capture-based control system to enable users to control the DigiArm using natural hand motions alone. Users wore reflective markers, and the system computed the required joint angles to effectively ``mimic'' their hand in real time.
The control software is programmed to recognize motion completion for specific joints, calculate motion velocity, and translate user intentions into standardized motor control messages (see Sec.~\ref{ssec:closed-loop_control}). Commands are transmitted via peer-to-peer WiFi to the hand controller, which parses the messages and thus actuates the motors.

\subsection{Key Press Force Characterization}
\label{ssec:key_press_force_requirements}


Prior to testing the hand with able-bodied users, we conducted an experiment to evaluate the fingertip force output as a function of applied voltage and characterize the key-pressing capabilities of our hand. Understanding the force generation capacity is critical for determining compatibility with standard keyboards and piano devices, which require specific actuation forces for reliable key pressing and force control.
The experimental setup (Fig.~\ref{fig:key_press_force_setup}) consisted of a single finger module mounted in a fixed position, a DC power supply, and a 6-axis force sensor positioned beneath the fingertip. We tested the system using different input voltage values, ranging from \SI{2}{\volt} to \SI{6}{\volt}, and measured the corresponding normal force during key-press motions.
Results (Fig.~\ref{fig:key_press_force_results}) demonstrate a progressive increase in force output with increasing input voltage, reaching approximately 3.6N at maximum voltage during key contact.



\subsection{User Study}
\label{ssec:user_study}


We collected data from eleven participants (6 male, 5 female) with an average age of $27$ years old. All participants were confirmed to be right-handed, experienced in playing piano, and without any neurological disorders or deformity in the hand. 
At the beginning of the experiment, participants were allowed to familiarize themselves with the software and robotic hand to ensure they could operate the system without issues. Each participant was instructed to perform two tasks: typing one short sentence (3-5 words) and playing one brief piano sequence (6-8 notes). Task lengths were intentionally kept short to minimize user fatigue, as operating the prosthetic hand in the ``mimicking'' setup required sustained arm elevation. Both tasks were performed under three conditions:
(1) with fingers restricted to the first level of the splay (closed and parallel fingers; see Fig.~\ref{fig:solid}, \textbf{C3}, right side),
(2) with the ability to choose and maintain a specific splay level for the recording period, and
(3) with full control, including finger movement, wrist motion, and adjustable splay positioning.
The six total sessions (two tasks $\times$ three conditions) were randomized to ensure that user performance was independent of learning effects. Participants were given at least one minute of rest between sessions to prevent fatigue. The study was approved by the institution's Ethics Committee and informed consent was obtained prior to data collection.



\textbf{What patterns emerge in finger usage during prosthetic-controlled typing tasks?}
We analyzed the distribution of finger usage across different key positions to understand how users adapt their typing strategies when controlling a robotic hand.
The results (see Fig.~\ref{fig:keyboard_heatmap}) indicate that users maintain systematic approaches to key allocation even when using a prosthetic hand.
The index finger was selected as the predominant digit for typing tasks, suggesting that users rely heavily on this finger for typing. In addition, analysis of spatial patterns shows increased utilization of the middle, ring, and little fingers for keys positioned toward the right side of the keyboard, indicating that users maintain some spatial mapping between finger position and key location.
Lastly, the thumb demonstrated clear dominance for space bar activation, with near-exclusive usage for this function across all users.
Meanwhile, while using the hand for piano playing (see Fig.~\ref{fig:piano_heatmap}), users were able to use the fingers in a relatively expected manner for accurate piano playing, specifically with the thumb and little finger used predominantly for notes further from the center, and the appropriate finger generally used for middle keys.

\begin{figure}[t]
\centering
\subfloat[\label{fig:subject_splay_decision}]{\includegraphics[trim=0 0 0 0, clip, width=0.16\textwidth]{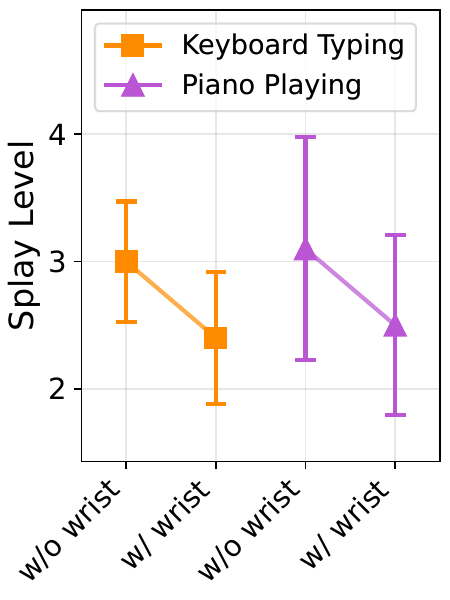}}\hspace{2.5mm}%
\subfloat[\label{fig:normalized_distances}]{\includegraphics[trim=0 0 0 0, clip, width=0.29\textwidth]{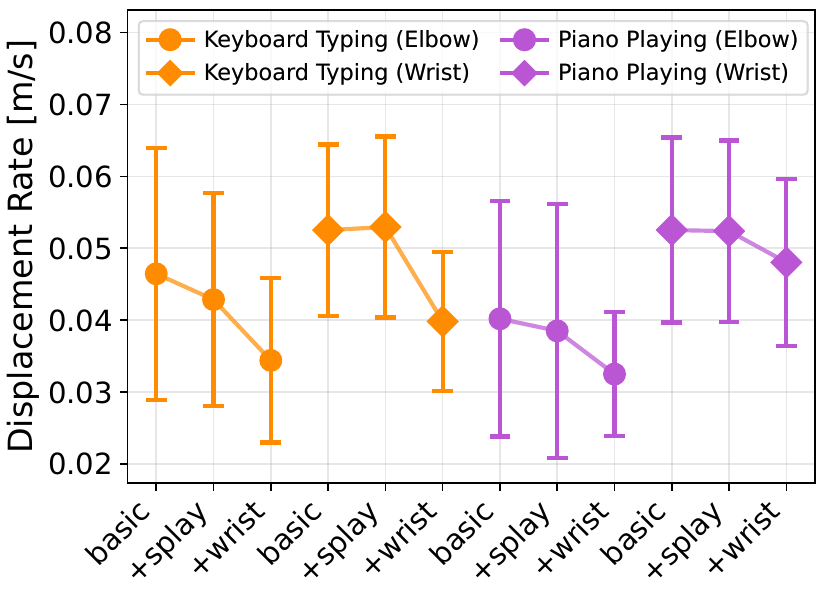}}
\vspace{-0.2cm}
\caption{
Impact of splay adjustment and wrist motion on user behavior and compensatory movements. 
(a) Average splay levels selected by users across experimental conditions, showing preference for increased finger spacing when wrist motion is restricted (level 1 = parallel fingers, level 5 = full spacing). Error bars represent standard error across eleven users.
(b) Elbow and wrist displacement rate during keyboard typing and piano playing tasks across three conditions: fixed parallel fingers (no splay, no wrist), adjustable finger spacing (with splay, no wrist), and full control (with splay and wrist). Results demonstrate progressive reduction in compensatory movements, with keyboard typing showing greater improvement than piano playing when both splay adjustment and wrist motion are available. Displacement rate is computed as distance traveled divided by key-pressing motion duration, averaged across all users.
}
\vspace{-0.5cm}
\end{figure}

\textbf{How do users adjust the splay mechanism when given the option to modify finger spacing?}
In two of the three conditions, users could select and adjust the splay level according to their preference for each task. The results (see Fig.~\ref{fig:subject_splay_decision}) demonstrate that users tend to select reduced splay levels when wrist motion is available compared to conditions with restricted wrist movement. Several participants reported preferring finger spacing similar to their natural hand configuration for typing tasks, but indicated they required less compensation when wrist control was accessible.
Additionally, we observed higher variance in splay level selection during piano playing tasks compared to typing tasks. This increased variability suggests that piano playing may require more finger spacing preferences, likely due to the varied chord structures and note combinations that benefit from different finger configurations, whereas typing presents more standardized spatial requirements.


\begin{figure}[t]
\centering
\includegraphics[trim=0 0 0 0, clip, width = 0.45\textwidth]{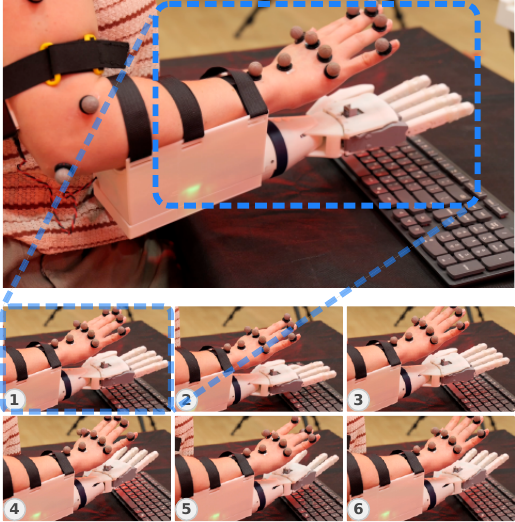}
\vspace{-0.15cm}
\caption{
Representative example of a user operating the DigiArm during keyboard typing.
The user starts with an open hand (1) and flexes the index finger to instruct the robotic hand to mimic the action and presses the keyboard (2). Then, the user lifts and extends the finger, followed by performing a wrist instruction to move the hand to the left (3 and 4). Finally, the user flexes the middle finger to press the keyboard once again and extend the finger back to its original position (5 and 6).
}
\label{fig:pressing_examples}
\vspace{-0.4cm}
\end{figure}


\textbf{How do splay adjustment and wrist motion contribute to reducing compensatory movements during key pressing?}
We measured the elbow and wrist displacement during key-pressing intervals for both piano playing and keyboard typing tasks using motion capture markers placed on the elbow and wrist joints (see Fig.~\ref{fig:pressing_examples}).
Measurements were taken as distance traveled in meters, normalized by the time interval of each key-pressing motion (defined as the interval from initial movement toward the target key until completion of key release), to evaluate whether enhanced finger spacing and wrist control reduce compensatory upper limb movements. The hypothesis was that providing users with adjustable finger spacing and wrist translation would decrease reliance on shoulder and elbow compensation strategies typically required with prostheses that lack these DOFs.

Results (see Fig.\ref{fig:normalized_distances}) demonstrate progressive reduction in compensatory movements across the three experimental conditions.
An initial improvement was observed when splay operation was allowed, followed by a notable reduction when wrist motion was enabled. 
Keyboard typing tasks showed slightly greater improvement compared to piano playing tasks, suggesting that typing applications may benefit more substantially from the enhanced DOFs. These findings indicate that enabling horizontal wrist motion and adjustable finger spacing substantially reduces the biomechanical effort required for digital interaction tasks, supporting our design for prosthetic hands intended for keyboard and piano applications.
Fig.~\ref{fig:pressing_examples} illustrates a representative example of a user operating the DigiArm during keyboard typing.

\subsection{Preserving Grasping Capabilities}
\label{ssec:preserving_grasping_capabilities}

\begin{figure}[t]
\centering
\subfloat{\includegraphics[trim=25 0 25 0, clip, width=0.155\textwidth]{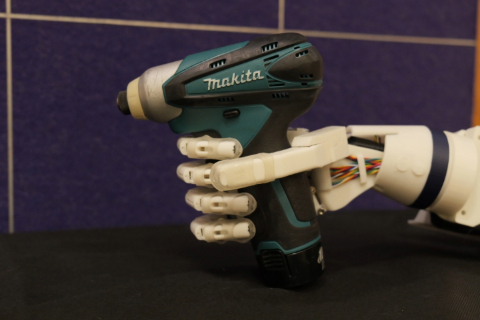}}\hspace{0.5mm}%
\subfloat{\includegraphics[trim=25 0 25 0, clip, width=0.155\textwidth]{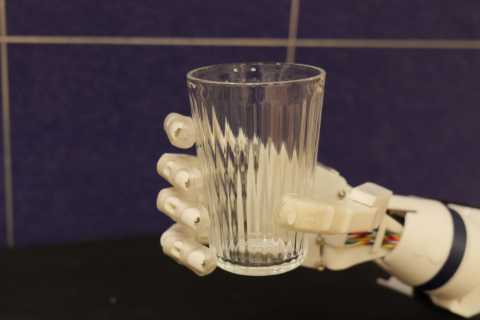}}\hspace{0.5mm}%
\subfloat{\includegraphics[trim=50 0 0 0, clip, width=0.155\textwidth]{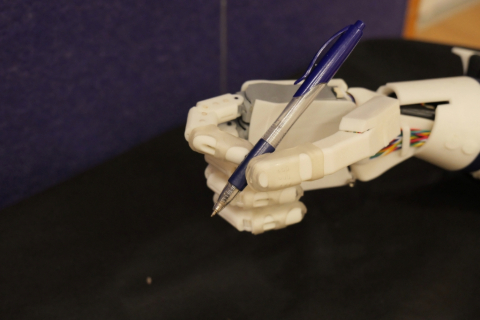}}\\ [-2.0ex]
\subfloat{\includegraphics[trim=25 0 25 0, clip, width=0.155\textwidth]{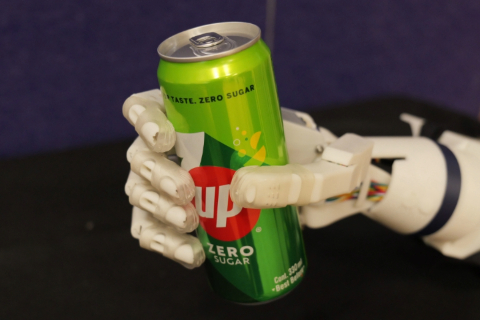}}\hspace{0.5mm}%
\subfloat{\includegraphics[trim=10 0 40 0, clip, width=0.155\textwidth]{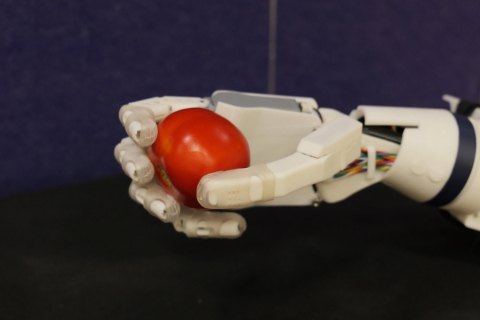}}\hspace{0.5mm}%
\subfloat{\includegraphics[trim=25 0 25 0, clip, width=0.155\textwidth]{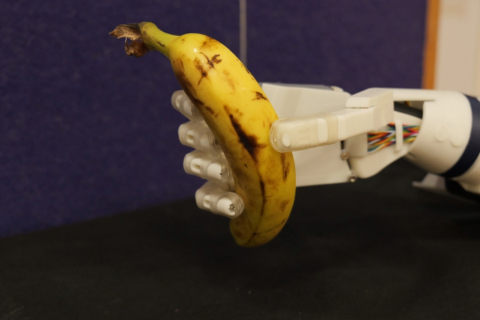}}
\vspace{-0.1cm}
\caption{
Grasping versatility for various everyday objects. The examples show successful grip adaptation for cylindrical objects (electric screwdriver, beverage can), a glass cup, a pen, a banana, and a tomato (spherical object). The hand accommodates different geometries in precision and power grips, maintaining a stable hold across varying materials, including metal, glass, and foods.}
\label{fig:objects_grasp}
\vspace{-0.3cm}
\end{figure}

To demonstrate that the DigiArm design preserves grasping capabilities alongside its key-pressing functions, we evaluated the ability of the hand to grasp and lift various everyday objects. We selected representative objects based on grasping assessments performed in previous prosthetic hands papers~\cite{controzzi2016sssa,weiner2018kit}.
The grasping protocol involved positioning the hand near each target object, followed by sequential actuation of the thumb and four fingers to secure the object. The DigiArm achieved 100\% success rate across all tested objects without any grasping failures. Representative examples presented in Fig.~\ref{fig:objects_grasp} demonstrate that the hand successfully grasps these objects and maintains stable grip strength sufficient for lifting heavier items, including an electric screwdriver weighing approximately \SI{2}{\kilo\gram}, despite being assembled only from 3D-printed components.


\section{CONCLUSIONS \& DISCUSSION}



This work presented the DigiArm, an open-source prosthetic hand platform specifically designed for key-pressing motions while preserving traditional grasping capabilities.
The $7$-DOF system presents novel features including the ``Splay'', a manual finger spacing adjustment, and powered ulnar/radial deviation, addressing critical gaps in current prosthetic hand functionality for keyboard typing and piano playing applications.
%
Our experimental validation demonstrates the design's effectiveness across multiple dimensions.
The user study with eleven users revealed progressive improvements as additional DOFs became available, with $19\%$ reduced compensatory elbow and shoulder movements when finger spacing adjustment and wrist motion were enabled.
Additionally, the preference for increased finger spacing when wrist motion was restricted confirms intuitive adaptation to compensate for missing DOFs, validating both the splay mechanism and ulnar/radial deviation control.
Finally, successful grasping trials with everyday objects up to \SI{2}{\kilo\gram} demonstrate that enhanced key-pressing functionality preserves traditional prosthetic capabilities.

Several constraints limit everyday adoption of the current design. The 3D-printed construction may lack durability for intensive use, while the fishing line system requires periodic maintenance that users may neglect. While the DigiArm alone is relatively lightweight at only \SI{341}{\gram}, the total system weight of \SI{648}{\gram} may cause fatigue during extended use. Finally, the manually adjusted splay, chosen to minimize actuator count, limits real-time adaptability during complex sequences.
Despite these limitations, the open-source design shows promise for clinical relevance with future modifications, while already serving as an experimental platform for prosthetics research.
The DigiArm demonstrates the benefits of added finger spacing and wrist deviation, establishing a foundation for next-generation prosthetic hands that will expand the range of accessible digital tasks available to upper-limb amputees.

\section*{ACKNOWLEDGMENT}

We thank the subjects who participated in the creation of the dataset.
We also thank Haifa3D, a non-profit organization providing self-made 3D-printed assistive solutions, for their support throughout this research.
We thank the Electronics and Software Lab for supplying the controller circuit used in this work.
This work was supported by the David Himelberg Foundation, the Wynn Family Foundation, the Israel Science Foundation (ISF) grant no.\ 1834/24, and the Israeli Ministry of Science \& Technology grants no.\ 7458 and 8477.





\section*{APPENDIX}
\label{sec:app}

\begin{figure}[t]
\centering
\includegraphics[trim=0 0 0 0, clip, width = 0.375\textwidth]{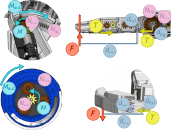}
\vspace{-0.1cm}
\caption{
Mechanical analysis diagrams.
(Left) Wrist torque analysis for ulnar/radial deviation and pronation/supination.
(Right) Force analysis for finger and thumb actuation.
}
\label{fig:forces_diagram}
\vspace{-0.1cm}
\end{figure}




\textbf{Ulnar/radial deviation torque analysis:}
The ulnar/radial deviation mechanism uses a bevel gear system where the motor (torque: $M = \SI{156.9}{\newton\milli\meter}$) drives a 12-tooth pinion ($N_{\text{pinion}}$) that meshes with a $132$-tooth large gear ($N_{\text{lg}}$) integrated into the bridge structure. The gear reduction provides a mechanical advantage of $11$:$1$, yielding the output moment
%
$$ M_{\text{out}} = M \frac{N_{\text{lg}}}{N_{\text{pinion}}} = \SI{156.9}{\newton\milli\meter} * \frac{132}{12} = \SI{1725.9}{\newton\milli\meter}$$

\textbf{Pronation/supination torque analysis:}
The pronation/supination motion employs a planetary gear system with the motor (torque: \SI{274.6}{\newton\milli\meter}) driving a $9$-tooth sun gear ($N_{\text{sun}}$). The system includes three $13$-tooth planet gears and a fixed $36$-tooth ring gear ($N_{\text{ring}}$). For this configuration, where the sun gear is the driving gear and the ring gear is fixed, the gear ratio is $1 + \frac{N_{\text{ring}}}{N_{\text{sun}}} = 1 + \frac{36}{9} = 5$. The output torque:
%
$$ M_{\text{out}} = M \Big(  1 + \frac{N_{\text{ring}}}{N_{\text{sun}}} \Big) = \SI{274.6}{\newton\milli\meter} * 5 = \SI{1373}{\newton\milli\meter}$$








\textbf{Single Finger Force Analysis:}
To estimate the fingertip force, we establish the following assumptions: 
(1) The gear train within the finger links constrains motion to a single DOF,
(2) The finger assembly behaves as a quasi-rigid body, making it possible to calculate the forces for every single point in time separately.
and (3) Internal gear train torques do not affect external force transmission.
The motor generates a torque of $M = \SI{100}{\newton\milli\meter}$ and drives a $6$-tooth pinion ($N_{\text{pinion}}$, the greyed pinion in Fig.~\ref{fig:forces_diagram}) that meshes with a $15$-tooth bevel gear ($N_{\text{bg}}$, the brown gear in Fig.~\ref{fig:forces_diagram}), providing a gear ratio of $2.5$:$1$. The bevel gear output torque is $M_{\text{bg}} = M\frac{N_{\text{bg}}}{N_{\text{pinion}}} = \SI{100}{\newton\milli\meter} * \frac{15}{6} = \SI{250}{\newton\milli\meter}$.
With the cable moment arm from the bevel gear center at $d_{\text{bg}} = \SI{5.5}{\milli\meter}$, the cable tension becomes: $T = \frac{M_{\text{bg}}}{d_{\text{bg}}} = \frac{250}{5.5} = \frac{500}{11}\SI{}{\newton} $.
Applying moment equilibrium about the finger joint ($F * d_{\text{length}} = T * d_{\text{cyl}}$) with $d_{\text{length}} = \SI{7.3}{\milli\meter}$ and $d_{\text{cyl}} = \SI{0.8}{\milli\meter}$:
%
$$ F = T\frac{d_{\text{cyl}}}{d_{\text{length}}} = \frac{500}{11}\SI{}{\newton} * \frac{0.8}{7.3} \sim \SI{4.98}{\newton}$$

\textbf{Thumb Force Analysis:}
The thumb motor produces $M = \SI{156.9}{\newton\milli\meter}$ and drives a spool mechanism with $d_{\text{spool}} = \SI{3.5}{\milli\meter}$, generating cable tension of $T =  \frac{M}{d_{\text{spool}}} = \frac{156.9}{3.5} \sim \SI{44.83}{\newton} $.
Using moment equilibrium with $d_{\text{offset}} = \SI{50}{\milli\meter}$ and  $d_{\text{cyl}} = \SI{3.5}{\milli\meter}$, we get:
%
$$ F = T\frac{d_{\text{cyl}}}{d_{\text{offset}}} = \SI{44.83}{\newton} * \frac{3.5}{50} \sim \SI{3.14}{\newton}$$

{\small
\bibliographystyle{IEEEtran}
\bibliography{main}
}

\end{document}